\documentclass[times,authoryear]{elsarticle}

\usepackage[utf8]{inputenc}
\usepackage{graphicx}
\usepackage{url}
\graphicspath{ {images/} }
\usepackage{caption}
\usepackage{amsmath}
\usepackage{verbatim}
\usepackage{subfig}

\usepackage{orcidlink}

\usepackage[authoryear]{natbib}

\RequirePackage{fix-cm}

\def\tsc#1{\csdef{#1}{\textsc{\lowercase{#1}}\xspace}}
\tsc{WGM}
\tsc{QE}
\tsc{EP}
\tsc{PMS}
\tsc{BEC}
\tsc{DE}

\usepackage{hyperref}
\hypersetup{
    colorlinks=true,
    linkcolor=blue,
    filecolor=magenta,      
    urlcolor=cyan,
    pdftitle={Overleaf Example},
    pdfpagemode=FullScreen,
    }    

\makeatletter
\def\ps@pprintTitle{%
  \let\@oddhead\@empty
  \let\@evenhead\@empty
  \let\@oddfoot\@empty
  \let\@evenfoot\@oddfoot
}
\makeatother

\begin{document}

\title{AI-driven multi-source data fusion for algal bloom severity classification in small inland water bodies: Leveraging Sentinel-2, DEM, and NOAA climate data}




\author[1]{Ioannis Nasios}

\ead{ioannis.nasios@nodalpoint.com}

\affiliation[1]{organization={Nodalpoint Systems},
    addressline={Pireos 205}, 
    city={Athens},
    postcode={118 53}, 
    country={Greece}}
    
\begin{frontmatter}

\begin{abstract}
Harmful algal blooms are a growing threat to inland water quality and public health worldwide, creating an urgent need for efficient, accurate, and cost-effective detection methods. This research introduces a high-performing methodology that integrates multiple open-source remote sensing data 
with advanced artificial intelligence models. Key data sources include Copernicus Sentinel-2 optical imagery, the Copernicus Digital Elevation Model (DEM), and NOAA’s High-Resolution Rapid Refresh (HRRR) climate data, all efficiently retrieved using platforms like Google Earth Engine (GEE) and Microsoft Planetary Computer (MPC). The NIR and two SWIR bands from Sentinel-2, the altitude from the elevation model, the temperature and wind from NOAA as well as the longitude and latitude were the most important features. The approach combines two types of machine learning models—tree-based models and a neural network—into an ensemble for classifying algal bloom severity. While the tree models performed strongly on their own, incorporating a neural network added robustness and demonstrated how deep learning models can effectively use diverse remote sensing inputs. The method leverages high-resolution satellite imagery and AI-driven analysis to monitor algal blooms dynamically, and although initially developed for a NASA competition in the U.S., it shows potential for global application. The complete code is available for further adaptation and practical implementation, illustrating the convergence of remote sensing data and AI to address critical environmental challenges ( \url{https://github.com/IoannisNasios/HarmfulAlgalBloomDetection}).
\end{abstract}


\end{frontmatter}

\section{Introduction}
\label{sec:introduction}

Algal blooms are becoming the greatest inland water quality threat to public health and aquatic ecosystems that can degrade water quality to a greater extent than many chemicals \citep{brooks2016harmful}. Climatic conditions is the main factor that drives the occurrence of blooms while nutrient enrichment is the basis for the growth and reproduction of cyanobacteria \citep{zhou2020analysis}. Human nutrient loading and climate change (warming, altered rainfall) synergistically enhance cyanobacterial blooms in aquatic ecosystems \citep{paerl2012climate}. Excessive nutrient loads in many cases comes from agricultural, industrial and other sources  \citep{novotny2011danger}. Phenology and trends of chlorophyll-a and cyanobacterial blooms are established \citep{matthews2014eutrophication}. To effectively reduce cyanobacterial biomass, and therefore limit health risks and frequencies of hypoxic events, a key underlying approach that should be considered in almost all instances is nutrient (both N and P) input reductions \citep{paerl2013harmful}. While minor health effects have been correlated with cyanobacterial contamination of drinking water drawn from rivers, the major unknown is the potential for cancer stimulation by cyanobacterial toxins, particularly gastrointestinal cancers in Australia and other affluent countries and liver cancer in poorer nations \citep{falconer2001toxic}. 

The application of fusing remote sensing data with other data sources for the detection of harmful algal blooms (HABs) holds substantial scientific importance, offering expansive coverage of water bodies and overcoming the limitations of conventional sampling methods. By providing synoptic and repetitive observations, remote sensing enables the identification of HAB occurrences on larger scales, facilitating early warning systems and informed decision-making. Moreover, it contributes to the formulation of predictive models for HAB dynamics, aiding in effective mitigation strategies, \cite{anderson2015living}. The non-invasive nature of remote sensing minimizes ecological disruption, while its ability to capture temporal trends facilitates the understanding of long-term HAB patterns and potential environmental linkages. In essence, remote sensing plays a pivotal role in comprehensively monitoring HABs, enhancing our capacity to safeguard aquatic ecosystems and human well-being from their detrimental impacts.

\begin{figure}[h]
    \centering
    \captionsetup{width=.5\linewidth}
    \includegraphics[width=0.5\textwidth]{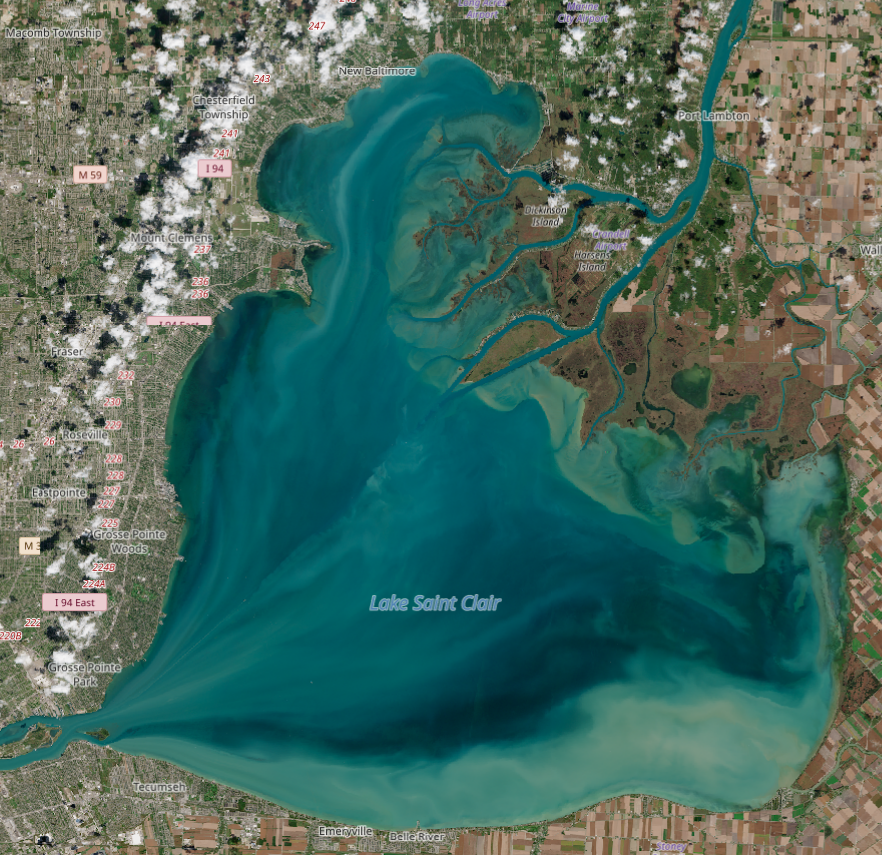}
    
    \caption{Algal bloom (light green) in Lake Saint Clair 
    (October 10, 2024). 
\href{https://browser.dataspace.copernicus.eu/?zoom=11&lat=42.51235&lng=-82.66457}{Source: Copernicus Sentinel-2}}

    \label{fig:StClaire101024}
    
\end{figure}

Inland water bodies are frequently situated within urban areas, where their ecological health is often compromised by human activities. \autoref{fig:StClaire101024} depicts an algal bloom in the southern section of Lake Saint Clair, near Detroit. The lighter green regions indicate areas with elevated algal concentrations, primarily attributed to nutrient inputs from river estuaries and agricultural runoff, which are likely the main drivers of this bloom.

This research comes after author's participation in a Machine Learning (ML) competition for detecting and classifying the severity of cyanobacterial algal blooms in small, inland water bodies, using various data sources. This challenge was created on behalf of \href{https://www.nasa.gov/}{NASA} with collaboration from \href{https://www.ncei.noaa.gov/}{NOAA}, \href{https://www.epa.gov/}{EPA}, \href{http://www.usgs.gov/}{USGS}, \href{https://www.diu.mil/}{DOD's Defense Innovation Unit}, \href{https://bair.berkeley.edu/}{Berkley AI Research}, and \href{https://www.microsoft.com/en-us/ai/ai-for-earth}{Microsoft AI for Earth}. The data for this challenge was a combination of satellite imagery, atmospheric model, elevation and tabular data as geolocation, date features and other. 

Open-source remote sensing data, such as Sentinel-2 imagery, can be accessed through the Copernicus Hub, either manually or via its API. This process, however, demands significant bandwidth, local disk space, and processing power. When applied to multiple images, these requirements increase both computational costs and processing time, particularly as each image typically spans extensive areas, while only a small section is often relevant for analysis. To address these challenges, platforms like Google Earth Engine (GEE) and Microsoft’s Planetary Computer (MPC) provide essential solutions, offering pre-processed satellite imagery where the specified Area of Interest (AOI) can be accessed directly, without the need to download and manually crop large datasets. GEE has become a widely used tool among researchers, as evidenced by the meta-analysis and systematic review conducted by \citet{tamiminia2020google}. MPC, while a newer platform, is emerging as an alternative. \citet{botelho2022mapping} used MPC to map roads in the Brazilian Amazon combining artificial intelligence and Sentinel-2, while \citet{xu2022analyzing} compared GEE and MPC for land cover mapping with deep learning in large-scale data cube analyses.

Numerous studies have leveraged remote sensing technology for detecting algal blooms across various aquatic environments. \citet{kislik2022mapping} used Sentinel-2 imagery within Google Earth Engine (GEE) to track algal bloom dynamics in small reservoirs, while \citet{alharbi2023remote} investigated algal blooms along the Red Sea Coast, identifying estuaries as key areas for algal concentration and growth. \citet{german2021space} found a strong correlation between algal bloom events and extreme chlorophyll-a levels in an Argentine eutrophic reservoir using Sentinel-2 data. \citet{seegers2021satellites} estimated chlorophyll-a and cyanobacteria indices across U.S. inland waters using MERIS time series data, while \citet{coffer2021satellite} examined cyanobacterial bloom frequency at multiple spatial scales through satellite remote sensing. \citet{cao2021spectral} developed a spectral index tailored to algal bloom detection with Sentinel-2 MSI imagery, and \citet{stumpf2016challenges} addressed the challenges of mapping cyanotoxin patterns from remote sensing data. \citet{ghatkar2019classification} classified algal bloom species using an extreme gradient boosted decision tree model. \citet{hunter2016remote} analyzed the effectiveness of various satellites in detecting cyanobacterial blooms in inland, coastal, and oceanic waters, and \citet{xu2021automatic} explored the potential of Sentinel-2 and Landsat imagery for temporal and spatial monitoring of algal blooms in Lake Taihu from 2017 to 2020.

Incorporating multiple data sources can significantly enhance model performance and facilitate a deeper understanding of area-specific conditions or research topics. \citet{arabi2020integration} combined in-situ and multi-sensor satellite observations for long-term monitoring of water quality in coastal areas. \citet{yang2020combined} leveraged the integration of Sentinel-2 and Landsat 8 data using Google Earth Engine (GEE) to monitor water surface area dynamics. Similarly, \citet{wang2021estimation} used stereo ZY-3, multispectral Sentinel-2, and DEM data to estimate tree height and above-ground biomass in North China's coniferous forests. \citet{zhang2022mapping} mapped irrigated croplands across China by employing a synergetic method for generating training samples, a machine learning classifier, and GEE. \citet{line2022using} utilized NOAA satellite imagery to detect and track hazardous sea spray in high-latitude regions and used Sentinel-2 data to manually create ground truth labels. This study further integrates diverse data sources to improve outcomes and provide essential insights for future research endeavors.

Remote sensing has emerged as an essential tool for water quality assessment and monitoring, offering extensive spatial coverage and timely data, which are crucial for effective resource management and public health protection. \citet{lioumbas2023satellite} monitored surface water quality in a large reservoir while \citet{barreneche2023monitoring} assessed different satellite image processing methods for chlorophyll-a estimation in Uruguay’s freshwater systems. \citet{keith2018monitoring} demonstrated the use of Landsat-8 imagery for monitoring algal blooms in drinking water reservoirs, highlighting the practical application of satellite data in water quality management.

This study addresses a critical gap by integrating multiple remote sensing data sources to enhance performance in algal bloom detection and monitoring while ensuring the approach remains efficient, fast, and cost-effective. By combining publicly accessible optical satellite imagery, satellite altimetry, and climate data, the methodology not only improves predictive performance but also emphasizes the significance of various environmental features contributing to algal bloom formation. Utilizing platforms such as Google Earth Engine, Microsoft’s Planetary Compute, and NOAA for data acquisition streamlines the process, making it accessible, cost-efficient, and comprehensible. This approach broadens usability beyond experts in earth observation, making it applicable and understandable to a wider audience, including resource managers and policymakers interested in environmental monitoring and protection.

\section{Material and methods}
\label{sec:Material and methods}
The methodology outlines the datasets, models, and complete workflow used for classifying the severity of cyanobacterial blooms in various inland water locations across the U.S. Each step is detailed, from the acquisition of remote sensing data to dataset construction, model training, prediction generation, ensemble modeling, and optimization of predictions to produce the final outputs.

\subsection{Data sources}
\label{sec:Data sources}
All data sources used in this study are publicly accessible, with the primary data source being the Copernicus Sentinel-2 imagery. In addition to Sentinel-2, Landsat-8 imagery was utilized as an available data source. Sentinel-2 imagery offers a higher spatial resolution in the visible spectrum (10 meters) than Landsat-8 (30 meters), although for data prior to mid-2015, only Landsat-8 imagery was available (data coverage extends back to 2013). Sentinel-2 and Landsat-8 satellites revisit the same locations approximately every five and eight days, respectively, providing imagery generally within a few days of in situ sampling events. Imagery within a ten-day range was considered representative of the sampling conditions. Additionally, high-resolution climatological data from the National Oceanic and Atmospheric Administration (NOAA) High-Resolution Rapid Refresh (HRRR) model, which provides hourly updated 3-km resolution data, was included. Finally, the 30-meter resolution Copernicus Digital Elevation Model (DEM) was incorporated, capturing Earth surface morphology, including buildings, infrastructure, and vegetation, based on satellite radar interferometry from the TanDEM-X Mission.

The four data sources that were used are:
\begin{description}
\item[$\bullet$]  Sentinel-2 satellite imagery 
\item[$\bullet$]  Landsat-8 satellite imagery 
\item[$\bullet$]  NOAA's High-Resolution Rapid Refresh (HRRR) atmospheric model's climate data 
\item[$\bullet$]  Copernicus DEM elevation data (collection cop-dem-glo-30)
\end{description}

Optical satellite imagery was accessed through Google Earth Engine (GEE) for this research, though preliminary experiments with Microsoft’s Planetary Computer (PC) yielded comparable results. HRRR data were sourced from the NOAA repository at https://noaahrrr.blob.core.windows.net/hrrr, while Copernicus DEM elevation data (30-meter resolution) were obtained via Microsoft’s Planetary Computer. All data retrieval processes were conducted using Python (for code availability, see \autoref{sec:Code availability}).

\subsection{Other data}
\label{sec:Other data}

\begin{figure}[h]
    \centering
    \includegraphics[width=0.4\textwidth,trim=0 0 0 23, clip]{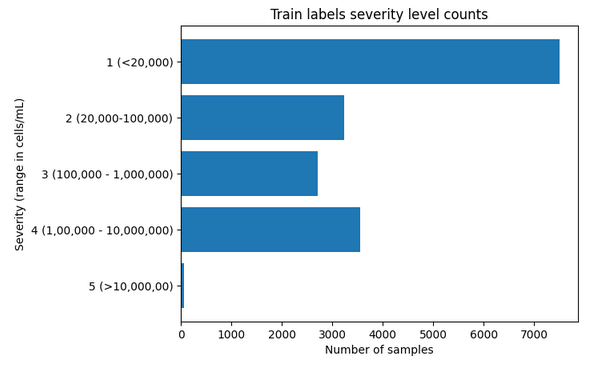}
    \caption{Severity histogram of train samples}
    \label{fig:severityhist}
\end{figure}

In addition to the satellite data sources, two supplementary CSV files were provided. The first file contained labels indicating the severity levels of samples along with their respective regions, while the second included geolocation data and sampling dates. These labels were based on in situ samples collected manually and subsequently analyzed in the lab to measure cyanobacteria density. Each sample represented a unique combination of date and location (latitude and longitude) and was sourced from inland water bodies across the continental U.S.. The goal was to predict the severity level (1, 2, 3, 4, or 5) based on cyanobacteria density. \autoref{fig:severityhist} provides a histogram of counts per severity level, alongside the associated cyanobacteria density. 

\begin{figure}[h]
    \centering
    \includegraphics[width=0.5\textwidth]{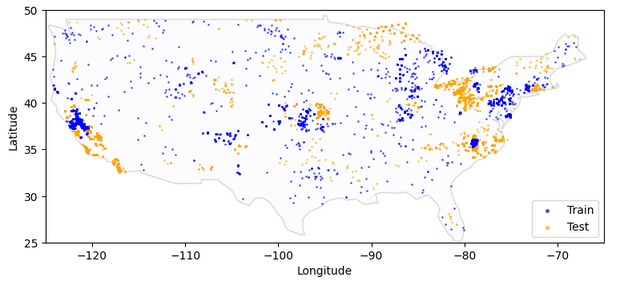}
    \caption{U.S.A map with train and test points of measurements}
    \label{fig:latlonpointsmap}
\end{figure}

In \autoref{fig:latlonpointsmap}, the sampling points are distributed throughout the continental United States. The training sample dates span from January 4, 2013, to December 14, 2021. The majority of the data is concentrated in the summer months, with fewer samples collected during the winter. This distribution reflects the increased risk of harmful algal blooms during the summer, when more individuals engage in recreational activities in water bodies such as lakes. Additionally, the higher temperatures and elevated human activity during the summer season contribute to a greater likelihood of algal blooms occurring during this time.

Different regions and data providers exhibit distinct patterns in the occurrence of harmful algal blooms. Some areas are represented by a high number of samples, which may dominate the overall dataset and disproportionately influence the trained models compared to regions with fewer samples. To address this imbalance and ensure a more generalized solution the sample weights were adjusted based on the region of origin. This approach ensures that samples from underrepresented regions contribute adequately to the model training process, thereby enhancing the model's ability to generalize across diverse geographical contexts.

In addition to the features obtained from the approved data sources, several critical features were derived from the metadata.csv file. These include geolocation data, specifically latitude and longitude, as well as temporal data such as month, year, and day of the week. Data exploration revealed a significant correlation between cyanobacteria severity and both latitude and longitude, highlighting the importance of sample location within the U.S. As a result, these features were incorporated into the model experiments, where they demonstrated substantial relevance to the predictive performance.

\subsection{Make datasets}
\label{sec:Make datasets}

Instead of selecting a narrow area around the sampling point, ideally the water area, from which water color features could be created, current methodology selects a wider area which most often includes land as well as water around the sampling points. The intuition behind this was that land usage such as crops, human constructions or other can affect the cyanobacteria concentration around the sampling area, \citep{mrdjen2018tile}. Furthermore, using this approach, shifted geolocation references, blocked from trees narrow canals or other similar obstacles wouldn't had high impact on performance. Therefore, the processing step of computational segmentation of the water area, which in some cases could had led to empty selected areas as segmentation wouldn't always be perfect, was omitted.

\begin{figure}[h]
    \centering
    \captionsetup{width=.5\linewidth}
    \includegraphics[width=0.5\textwidth]{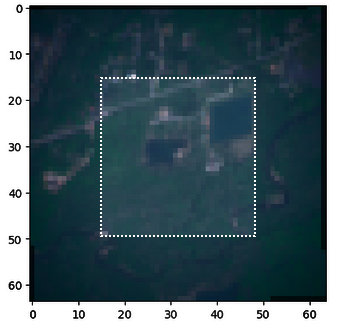}
    \caption{Sentinel-2 RGB image 64x64 and center cropped in 32x32 pixels}
    \label{fig:RGB64hl32stroke}
\end{figure}

The metadata file contained a total of 23570 samples, which included both training and test datasets (leaderboard). For each sample, data from the approved sources were requested. The downloaded Sentinel-2 data generated a dataset comprising 23130 images, structured as (23130, 64, 64, 7), while the Landsat-8 dataset included 20035 images, formatted as (20035, 64, 64, 7). Each image was initially scaled to the range [0-1], then multiplied by 255, and converted to the uint8 data type. The datasets consisted of 7 channels: for Sentinel-2, these were B1, B2, B3, B4, B8, B11, and B12, while for Landsat-8, the equivalent wavelengths were B1, B2, B3, B4, B5, B6, and B7. The initially retrieved images represented an area of 500 meters, with the point of interest centered (250 meters on each side). Since Google Earth Engine (GEE) queries are based on specified area size, the returned images may vary slightly in pixel dimensions depending on the location. To create a uniform dataset, all images were resized to 64x64 pixels. An important parameter was the time window for the imagery, defined as the maximum duration of 20 days before the sampling date. Additionally, acquired images were required to have less than 30\% cloud cover. These limitations along with the revisit times resulted the corresponding datasets to luck imagery in many cases, especially for the Landsat satellite. Afterwards, the datasets were center-cropped to a shape of (number of images, 32, 32, 7), as illustrated in the sample image of \autoref{fig:RGB64hl32stroke}, reducing the area around the sampling point to approximately 250x250 meters. It was found that retrieving images at a size of 64x64 pixels and then cropping to 32x32 pixels produced slightly better results compared to obtaining images directly at 32x32 pixels. This is likely due to improved scaling, as image processing algorithms utilize the minimum and maximum values for scaling, with larger images providing better statistical information. The processed images were then prepared for input into the neural networks, while for tree-based models, statistical features were calculated using the mean, median, and standard deviation of pixel values for each channel in every image.

The "COPERNICUS/S2\_HARMONIZED" GEE image collection for sentinel-2 images and the "LANDSAT/LC08/C01/T1\_SR" for Landsat-8 images were retrieved. A comparison between spectral characteristics of Landsat-8 and Sentinel-2 can be seen in \autoref{table:LS8S2}.

\begin{table*}[ht!]
\caption{Landsat-8 vs Sentinel-2.}
\label{table:LS8S2}
\centering
\begin{tabular}{l  c c c c c c } 
    &   &        LANDSAT-8    & &   &      SENTINEL-2  \\
 \hline
 Band name & Band & Wavelength & Res. & Band & Wavelength & Res. \\
 \hline

Coastal Aerosol & 1 & 0.45–0.451  & 30 & 1  & 0.430–0.450 & 60 \\

Blue            & 2 & 0.452–0.51  & 30 & 2  & 0.448–0.546 & 10 \\

Green           & 3 & 0.533–0.59  & 30 & 3  & 0.538–0.583 & 10 \\

Red             & 4 & 0.636–0.673 & 30 & 4  & 0.646–0.684 & 10 \\

NIR             & 5 & 0.851–0.879 & 30 & 8A & 0.848–0.881 & 20 \\

NIR2            &   &             &    & 8 & 0.763–0.908 & 10 \\

SWIR1           & 6 & 1.566–1.651 & 30 & 11 & 1.542–1.685 & 20 \\

SWIR2           & 7 & 2.107–2.294 & 30 & 12 & 2.081–2.323 & 20 \\

Cirrus          & 9 & 1.363–1.384 & 30 & 10 & 1.336–1.441 & 60 \\


\multicolumn{5}{l}\footnotesize{Resolution (Res.) in meters, wavelength in micrometers}\\

\end{tabular}
\end{table*}

\begin{figure*}[h]
    \centering
    \includegraphics[width=0.75\textwidth]{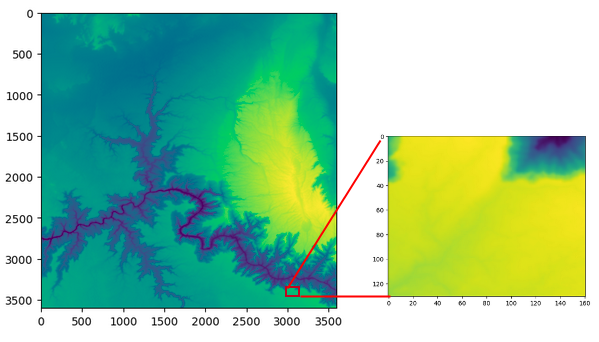}
    \caption{Planetary Computer DEM tile and cropped area around our point of interest}
    \label{fig:DEMtileandcrop}
\end{figure*}

The Copernicus Digital Elevation Model (DEM) provides elevation data at a resolution of 30 meters. This type of data has been shown to enhance classification performance when fused with Sentinel-2 imagery in other studies \citep{hosseiny2022urban}. While elevation data may initially appear irrelevant for detecting harmful algal blooms (HABs), characteristics such as altitude and geomorphology can significantly improve overall modeling performance. At the Planetary Computer, this data is stored in tiles measuring 3600x3600 pixels, covering an area of approximately 108x108 kilometers, depending on the location. To construct the necessary dataset, an area of 4000x4000 meters 
surrounding each sampling point was initially retrieved, as illustrated in \autoref{fig:DEMtileandcrop}. These images were subsequently resized to 32x32 pixels to reduce their size for convenience, with minimal loss of information during the resizing process. Each image was scaled and converted to the uint8 data type, similar to the satellite images described previously. This scaling process removes absolute altitude information from the elevation images, even though altitude is an important feature. Therefore, a separate feature vector containing the absolute altitude at each sampling point was saved.
Additionally, the elevation images were transformed into geomorphological images, enabling geomorphological comparisons between samples. As a result, a geomorphological dataset with the shape (23556, 32, 32) was constructed, along with a corresponding feature vector of length 23556.

\begin{figure*}[h]
    \centering
    \includegraphics[width=0.99\textwidth]{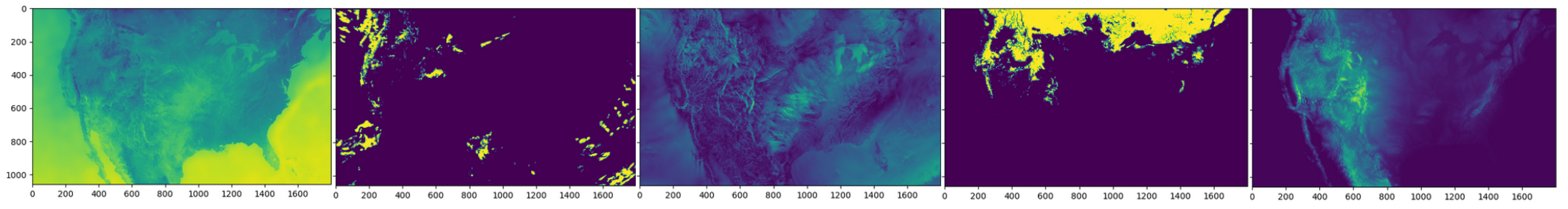}
    \caption{Date 2020 11 19. USA maps of Temperature, Rain, Gust, Snowc and Hgt.}
    \label{fig:NOAA5}
\end{figure*}

Out of many available climatological features, only temperature, rain, gust (wind sudden increase), snow and HGT (Geopotential Heights on Pressure Levels) were used. For every sample point and time, the past 28 days of selected climatological raw features were retrieved forming a dataset of shape (20502 x 28 x 5). There were several data points missing from this dataset, as it was not able to be retrieved. From this dataset, statistical features was made for the tree models, with the mean and std values of every channel of every sample. Neural network submodel was fed with the whole timeseries using a recurrent neural network submodel. Thumbnail maps strip in \autoref{fig:NOAA5} are of a sample date, for the whole continental USA.

\subsection{Metric}
\label{sec:Metric}
\begin{figure}[h]
    \centering
    \includegraphics[width=0.5\textwidth,trim=0 0 0 23, clip]{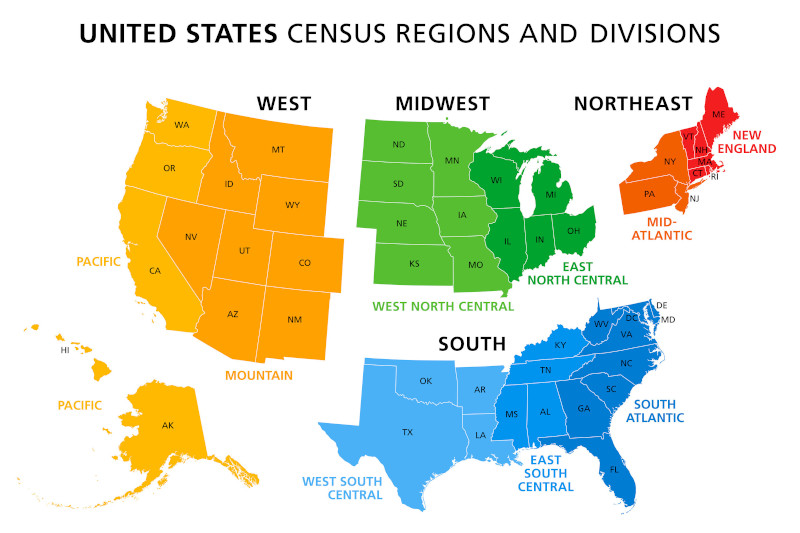}
    \caption{USA map of regions. West – MidWest – South – NorthEast }
    \label{fig:competition_cyano_us_regions2}
\end{figure}

To make sure models were incentivized to perform well across the U.S., and not just on the most represented areas, RMSE was calculated separately for each region (West, Midwest, South, and Northeast) and then averaged. A U.S. map divided to the regions of interest can be seen in \autoref{fig:competition_cyano_us_regions2}.

Competition's metric was the Region-Averaged Root Mean Squared Error (RA-RMSE). RMSE is the square root of the mean of squared differences between estimated and observed values. As this is an error metric, a lower value is better. 

\begin{equation}
\displaystyle
RMSE = \sqrt{ \frac{1}{N}  \sum\limits_{i=0}^{N} (y_{i} - \hat{y_{i}} )^2  }
\label{eqn:rmse}
\end{equation}

In \autoref{eqn:rmse}, \(\hat{y_{i}}\) is the predicted severity level for the \(i_{th}\) sample, \(y_{i}\) is the actual severity level for the \(i_{th}\) sample and N is the number of samples.

RMSE was calculated separately for each one of the four regions of the U.S. 
The final metric was the average of all region-specific RMSEs of \autoref{eqn:ra-rmse}. 

\begin{equation}
\displaystyle
Final score = \frac{West + Midwest + South + Northeast}{4}
\label{eqn:ra-rmse}
\end{equation}

\subsection{Modelling}
\label{sec:Modelling}
Presented modelling methodology here is a simpler version, ranked 6th instead of author's best 5th place. This method scored 0.8110 on the private LB as trading a small performance loss with a more friendly solution more convenient and simpler was preferred. The differences between presented here and best approaches include multiple trainings with various target transformations for the best, usage of only the first three of the climatological data, training an additional lightgbm model with all features and other minor modifications. 

Three machine learning models were trained, 1 random forest \citep{breiman2001random}, 1 lightgbm \citep{ke2017lightgbm} and 1 keras over tensorflow neural network. The lightgbm model trained using only the most important features of the random forest model. Instead of severity, square root transformation of density was used as target for model training (target transformation). Finally, the ensembled output was converted to severity by utilizing an optimization function, based on scipy's library optimize.minimize function with Nelder-Mead method \citep{gao2012implementing}, tuned to the Out of Fold (OOF) predictions. The validation scheme that was used is the 5-fold cross validation. For every fold and every model, predictions on the out of fold part of training data was saved along with predictions on test data, for later use by the optimization function.


As tree models are very different from 2D CNN models regarding the shape of data (1D instead of 2D), 2 different datasets were created. The random forest model used the following 45 features.

\begin{description}
\item[$\bullet$]  \textbf{channel statistical features} containing for every sentinel-2 patch, the mean, median and standard deviation of every channel (3*7=21 total features). 
\item[$\bullet$]  \textbf{B3B2}, (patch\_medianB3-patch\_medianB2)/(patch\_medianB3+patch\_medianB2)
\item[$\bullet$]  \textbf{B3B4}, (patch\_medianB3-patch\_medianB4)/(patch\_medianB3+patch\_medianB4)
\item[$\bullet$]  \textbf{B5B4}, (patch\_medianB5-patch\_medianB4)/(patch\_medianB5+patch\_medianB4)
\item[$\bullet$]  \textbf{date features}: month, year and dayofweek
\item[$\bullet$]  \textbf{location features}: latitude and longitude
\item[$\bullet$]  \textbf{climatological features}: temperature mean, temperature std, temperature mean fw, temperature std fw, rain mean, rain std, gust mean, gust std, snowc mean, snowc std, hgt mean and hgt std. (fw is from 1 week before sampling date whether all other stats are from 4 weeks prior sampling, 12 total features)
\item[$\bullet$]  \textbf{geomorphological features}: altitude, DEMmean, DEMmedian and DEMstd. (DEM is the geomorphology image around our sampling point)
\end{description}

Null or infinite values were replaced by -999. The Random forest model trained with n\_estimators equal to 300 while used default values for all other hyperparameters. For the lightgbm model, only the 19 most important features (\autoref{fig:lgbm19}), derived from random forest feature importance were used (where importance $>$0.005). This feature importance derived as the average across all folds (5 folds). The Lightgbm model was trained without early stopping, for 600 rounds and with a learning rate equal to 0.025. Other important hyperparameters different from default are: bagging\_fraction=0.9, bagging\_freq=6, subsample=0.8, and num\_leaves=20. Finally, for training sample weights were used based on sample region (midwest=1.51, northeast=1.98, south=1.11 and west=1.3). These weights derived after experimentation 
for increasing model's performance. 
    
\begin{figure*}[h]
    \centering
    \includegraphics[width=0.8\textwidth]{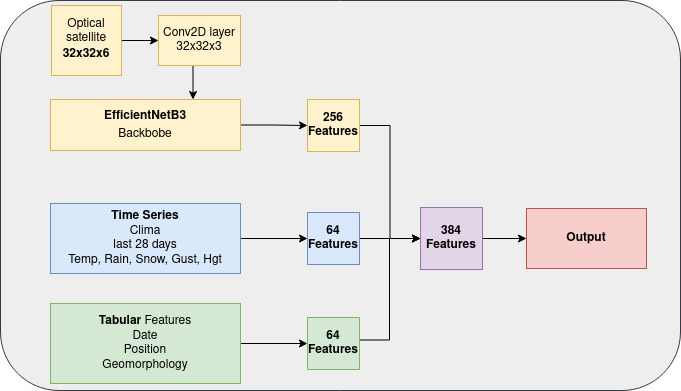}
    \caption{ Data fusion within Neural network }
    \label{fig:NN_basic_arch}
\end{figure*}

The neural network comprises three distinct submodels that process different types of input data, as shown in \autoref{fig:NN_basic_arch}. The first submodel handles optical satellite data, using only the last six of the seven available channels. The coastal aerosol band was excluded as this improved model's performance. Later the six channels are reduced to three with a convolutional layer in order to pass through the EfficientNetB3 backbone \citep{tan2019efficientnet}. The second submodel is a recurrent neural network designed for climate features, processing five weather variables—temperature, rainfall, gust speed, snow cover, and geopotential height—over 28 days, with each feature individually standard-scaled. The third submodel processes tabular data, including date features, 
geolocation features (latitude and longitude), and geomorphological features (altitude, DEM standard deviation, DEM mean, and DEM median), all of which are also standardized. The outputs from these submodels are concatenated, yielding 256 features from the satellite submodel, 64 from the climate submodel, and 64 from the tabular submodel, resulting in a total of 384 high-level features passed to the output layer.

Neural network training characteristics include the sample weights per region, to deal with the imbalance of number of samples between regions. Sample region weights used are: midwest=0.86, northeast=1.12, south=0.58 and west=0.71. As in the lightGBM model, also for the neural network, sample weights derived after experimentation 
for increasing model's performance. Also, the cosine annealing learning rate schedule with total 60 epochs in 2 snapshots, starting from learning rate equal to 0.001 and the Adam optimizer were used. Furthermore, the image augmentation that was used, includes horizontal and vertical flips, width and height shift (0.1) and zoom (0.2). Final, for inference all 4 possible estimations using flips of the satellite image were averaged (test time augmentation, TTA).

\begin{figure*}[h]
    \centering
    \includegraphics[width=0.99\textwidth]{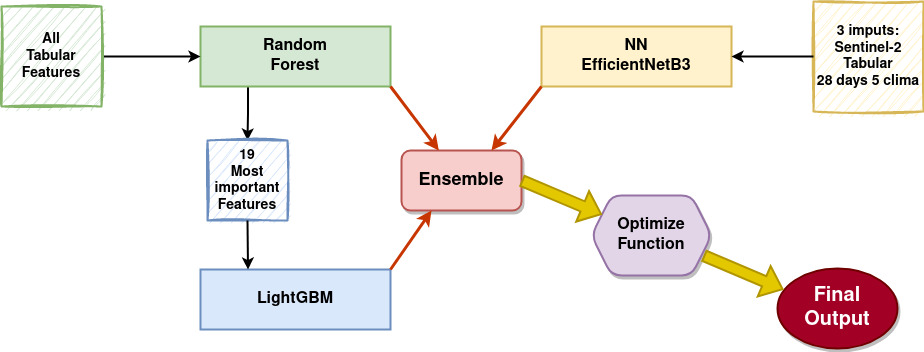}
    \caption{Fusing model predictions - Flowchart }
    \label{fig:flowchart_HAB}
\end{figure*}

\begin{figure}[h]
    \centering
    \captionsetup{width=.5\linewidth}
    \includegraphics[width=0.45\textwidth]{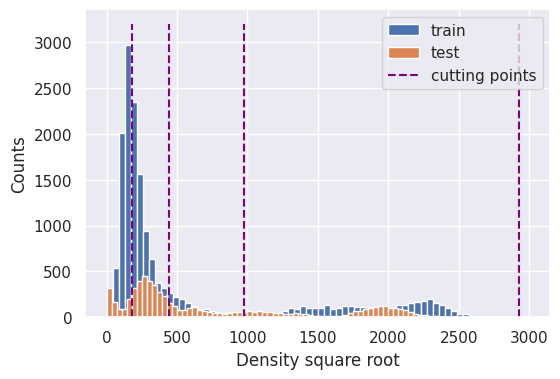}
    \caption{Histogram of OOF train and test predictions with cutting points}
    \label{fig:densityhistswithcoefs}
\end{figure}

Flowchart in \autoref{fig:flowchart_HAB} shows the whole process from datasets to the final outcome. All 3 models estimations are averaged to a fused prediction. As models trained with the square root of cyanobacterial density as the target, predicted estimations are also in this range. To transform these predictions to severity, a custom optimization function is used which was tuned to the OOF predictions to find the optimal cutting points for classes separation. \autoref{fig:densityhistswithcoefs} shows the histogram of the OOF predictions, a histogram with test predictions and the cutting points, the class separation points used to convert the square root of cyanobacteria density to severity. These cutting points are [180, 440, 979 and 2926], estimations of less than 180 are of severity equal to 1, 180-440 equals 2, 440-979 equals 3, 979-2926 equals 4 and greater than 2926 equals 5. Finally, estimations are clipped to 4 as a maximum predicted value.

\section{Results and discussion}
\label{sec:Results and discussion}
The results from the competition’s private leaderboard (LB) are presented in \autoref{table:competitions_results}. Most solutions significantly outperformed the benchmark, with the top solution cutting the error in half. A key advantage of the approach presented here is its integration of both tree models and neural networks, a feature not shared by other top-performing solutions. The effective combination of deep learning models with multiple remote sensing data sources demonstrated in this work is noteworthy, as it could also offer valuable insights for other applications that require integrating diverse data types. Furthermore, while the proposed methodology scored slightly lower than the winning solutions in this specific competition, it has the potential to perform better in different regions, on alternate test datasets, or for specific areas of interest.

\begin{table}[ht!]
\caption{Competitions results, author's performance in bold}
\label{table:competitions_results}
\centering
\begin{tabular}{l  l  c } 
 \hline
 Rank & Participant & RA-RMSE error \\ [1ex] 
 \hline
 1 & sheep & 0.7608  \\ 
 2 & apwheele & 0.7616  \\
 3 & karelds & 0.7844  \\
 4 & BrandenKMurray & 0.7879  \\ 
 5 & \textbf{ouranos} & \textbf{0.8034} \\ 
 
 \hline
 Benchmark & &  1.5724 \\ 
 \hline
\end{tabular}

\end{table}

None of the winning solutions employed neural networks but instead, all relied on gradient-boosted tree models. All generated overall image statistics, such as the ratio of blue to green bands and the mean values of each band. Techniques to identify water bodies within the images were also developed, either through K-means segmentation or by utilizing the scene classification band from Sentinel-2 imagery. Additionally, all solutions incorporated environmental features from external sources, predominantly using temperature, humidity, and elevation, while other variables like precipitation and wind were ultimately deemed unhelpful for modeling purposes. The first-place winner highlighted the South region as particularly problematic due to inaccurate location data, labeling it as noisy, and noted the difficulty of predicting outcomes in the West region, where many samples came from small rivers less than 10 meters wide, a challenge given Sentinel-2’s resolution. Further details on the winning solutions can be found at the \href{https://drivendata.co/blog/tick-tick-bloom-challenge-winners}{competition's winners report}.

\autoref{table:oof_results} shows the OOF error for the RF, LightGBM and NN models per region, as well as their ensemble score (simple predictions average). NN performed best in the South and RF at Northeast and Midwest while the ensemble solution worked best in all regions average.

\begin{table*}[ht!]
\caption{OOF scores}
\label{table:oof_results}
\centering
\begin{tabular}{l  c c c c c c } 
 \hline
 model & South & West & Northeast & Midwest & Region mean & samples mean \\ [1ex] 
 \hline
 RF & 0.769 &  0.424 & \textbf{0.818} & \textbf{0.809} & 0.705 & 0.717 \\ 
 LGBM & 0.801 &  0.419 & 0.866 & 0.87 & 0.739 & 0.749 \\  
 NN & \textbf{0.761} &  0.415 & 0.874 & 0.865 & 0.749 & 0.773 \\ 
 \hline
 Ensemble & 0.763 &  \textbf{0.398} & 0.822 & 0.816 & \textbf{0.700} & \textbf{0.711} \\ 
 \hline
\end{tabular}

\end{table*}

Segmenting water region in satellite images and creating color features was the key element to win this competition. Although our method did not crop out land areas from the satellite images, it still produced highly competitive results. In the NorthEast and MidWest regions, the area around the sampling point, including crops, buildings or any other structures were not as important as features derived from various bands of the Sentinel-2 satellite in the water area. Additionally, the superior performance of neural networks compared to tree models in the South—an area plagued by inaccurate sampling locations—and in the West—characterized by narrow canals—demonstrated the effectiveness of using a larger contextual area around each measurement point.

In the confusion matrix of \autoref{table:oof_cm} the OOF final predictions are compared with the true severity labels. As expected, most of the miss-classifications were classified as their neighbor class. In terms of accuracy, the overall accuracy was about 62.8\% (10720/17060) while for the most severe cases with true label greater or equal to 4 there was an accuracy of about 89.5\% (3228/3605).  

\begin{table}[ht!]
\caption{OOF confusion matrix}
\label{table:oof_cm}
\centering
\begin{tabular}{l  c c c c c } 
 \hline
   & pred 1 & pred 2 & pred 3 & pred 4 & pred 5 \\ [1ex] 
 \hline
 true 1 & 4975 &  2347 & 152 & 22   & 1 \\  
 true 2 & 1096 &  1825 & 275 & 37   & 6 \\  
 true 3 & 343  &  1382 & 692 & 300  & 2 \\  
 true 4 & 7    &  106  & 205 & 3226 & 3 \\  
 true 5 & 0    &  8    & 19  & 29   & 2 \\  
 
\end{tabular}

\end{table}

\begin{figure*}[h]
    \centering
    \includegraphics[width=0.8\textwidth]{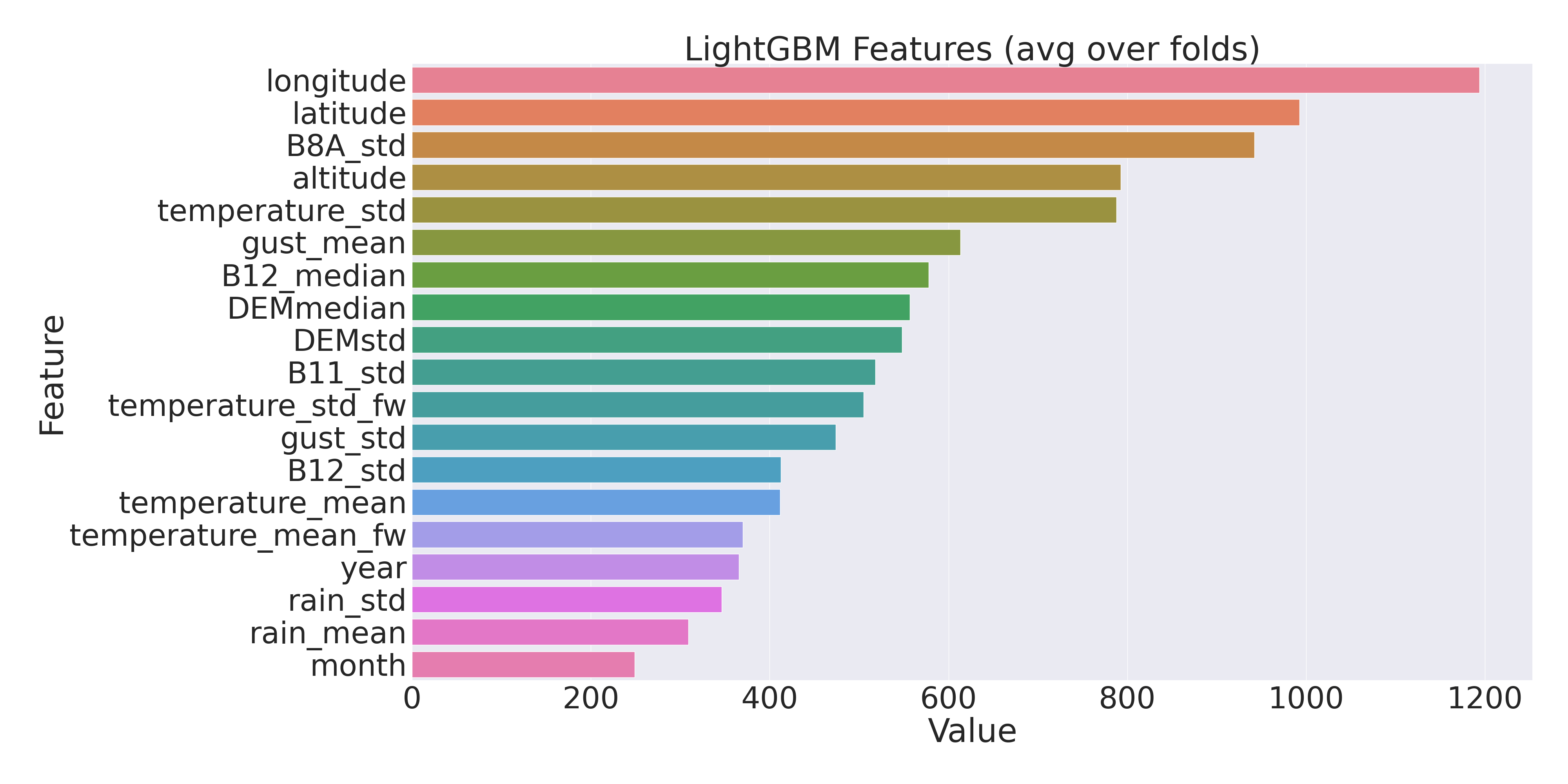}
    \caption{LightGBM feature importance }
    \label{fig:lgbm19}
\end{figure*}

Longitude was by far the most important feature while climatological features were more important than optical satellite features for the random forest model. For the lightGBM model, longitude was also the most important feature but not as important as for the RF model while features from all sources substantially contributed. The NIR (B8A) and the 2 SWIR bands (B11 and B12) were the most important from all the Sentinel-2, their importance was greater than the Red, Green or Blue higher resolution bands. As the value of NIR and SWIR bands for vegetation is well studied in remote sensing this was expected and the rule was verified. The altitude was the most important from the DEM data contributing to increase method's performance. Finally, temperature, both over 28 days period and 7-days, and wind (28d) were the most important from all the climatological features. \autoref{fig:lgbm19} shows the 19 most important features as derived from the RF model and their importance for the lightGBM model.

\begin{figure}[h]
    \centering
    \captionsetup{width=.9\linewidth}
    \includegraphics[width=0.5\textwidth]{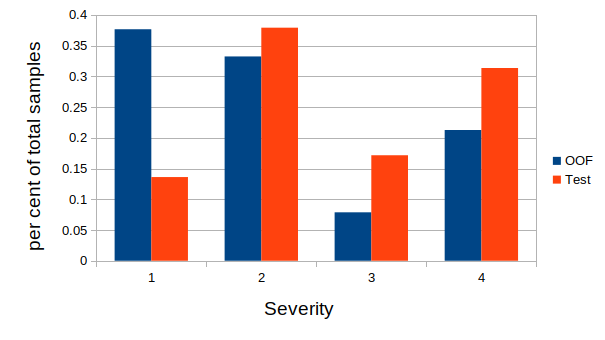}
    \caption{OOF vs Test final predictions }
    \label{fig:oof_test_predictions_severity_hist}
\end{figure}

Test samples distribution as can be seen in combined histogram of severity predictions of \autoref{fig:oof_test_predictions_severity_hist} is quite different from train samples. Samples in test data are in general of higher cyanobacteria severity. This is also reflected partially in the difference between the OOF score and the LB score as the LB score was worse. This difference in distributions between the train and test dataset was an extra issue that had to be addressed and more attention was given for trained models not to be overconfident.

All 3 different sources of data contributed to increase modelling performance. Sentinel-2 optical satellite, HRRR climatological data and DEM elevation data, provided valuable features to the models. Experiments conducted without optical satellite or climate or elevation data performed worse. Also, experiments made with neural networks using geomorphology data as images and not as statistical features, as well as using climatological data statistical features instead of timeseries, also performed worse.

\begin{figure}[h]
    \centering
    \subfloat[\centering Sentinel-2]
    {{\includegraphics[width=5cm,trim=0 0 0 0, clip]{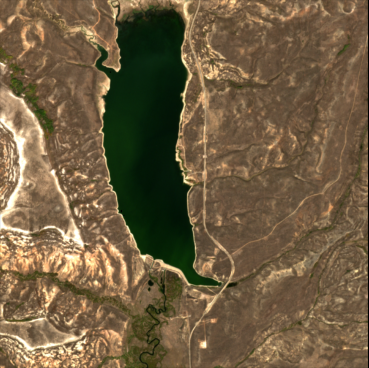} }}%
    \subfloat[\centering Landsat-8]
    {{\includegraphics[width=5cm,trim=0 0 0 0, clip]{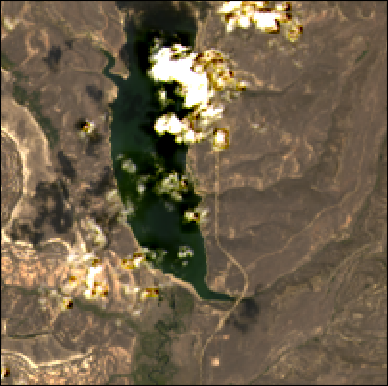} }}%
    \caption{6x6 KM of Lake Viva Naughton in Wyoming. }%
    \label{fig:S0vsL0}%
\end{figure}

Despite conducting numerous experiments with Landsat-8 data, these data were not utilized in the final solution. Only Sentinel-2 data were used as the source of optical satellite information. The superior performance observed with Sentinel-2 data, compared to Landsat-8, was likely due to the higher spatial resolution of Sentinel-2 satellites (\autoref{fig:S0vsL0}). Additionally, the greater amount of missing data in the Landsat dataset may have further reduced its effectiveness. For dates prior to mid-2015 or in cases where no Sentinel-2 image was available, synthetic images were generated by setting all pixels to the mean pixel value of existing Sentinel-2 images for each channel.

Using GEE and MPC as data sources was easy and data download was quick and already processed accelerating the machine learning experimentation. Furthermore, resources as the network bandwidth, CPU and RAM used for dataset creation were minimum.


Using both tree models and Neural Networks with multiple remote sensing inputs increased performance. Including pretrained neural networks for the optical satellite images contributed to the overall solution's performance and robustness. Showcasing how to use neural networks with multiple remote sensing data effectively, can provide insights to future projects, research and applications. Tree models with their speed and ability to run fast on CPU can be used for inference very easy and efficient alone for this task with high accuracy.

The HRRR dataset provides numerous climatological features, but for this approach, only temperature, rain, gust (wind), snow, and geopotential height (Hgt) were initially selected. Further experimentation indicated that using only temperature, rain, and wind yielded effective results. Since temperature plays a critical role in triggering algal blooms, climate data were collected not only for the sampling dates but also over the preceding four weeks. This extended period was chosen because weather conditions on the sampling day might not fully represent the conditions leading up to an algal bloom, as blooms can persist for several days once initiated. Favorable weather patterns from previous days may have set off the bloom-formation mechanisms. While the choice of a four-week window could be subject to further investigation, any variations in the window length are unlikely to produce significantly different outcomes, particularly given the presence of other contributing data types.

Fusion has played a crucial role in this research, occurring frequently and in various forms. The most evident example is the integration of diverse data sources, including Sentinel-2 imagery, DEM elevation data, climate information from atmospheric models, historical datasets, as well as the geolocation and date of each sampling point. Fusion also occurred at the model level, where different model predictions were combined through ensembling. Additionally, since satellite data include numerous bands beyond just the three RGB channels typical of standard images, utilizing all these bands represents a form of image-level fusion. Lastly, using a dataset that spans diverse locations across the U.S. enabled the fusion of comprehensive and representative information into the machine learning models. This broad geographic coverage provided a robust basis for training models that, despite being developed with nationwide data, can effectively be applied to make inferences in specific subregions.

\subsection{Perspectives}
\label{sec:Perspectives}
Sentinel-2 and Landsat-8 imagery, along with Copernicus DEM elevation data, offer global coverage. In contrast, NOAA's climate data are limited to the U.S., which poses a challenge for developing machine learning models intended for use outside the United States. To create globally applicable models, NOAA's data and associated metadata would need to be excluded. Additionally, the geolocation features that are influential within the U.S. may not have the same impact in other regions. Although the provided labels for U.S. inland water bodies can be used to train models capable of performing inference globally, the reliability and uncertainty of such predictions would increase when applied outside the U.S. Expanding the current dataset to include similar labels from other regions worldwide would significantly enhance the development of a robust global model. Future experiments could also explore additional data sources, such as another Copernicus Sentinel satellite, which might improve performance both locally and on a global scale.

Given the regular updates to remote sensing and atmospheric model data, the current methodology could serve as a foundational component for a monitoring system. This system could periodically execute the inference pipeline using the latest data to generate a dynamic live map of cyanobacteria severity for all inland water bodies within a designated Area of Interest (AOI), whether that be within a city or across the entire U.S. High-severity cyanobacteria estimates could trigger alerts to warn about potential safety hazards, and the enhanced accuracy of the methodology, particularly for severe cases, supports this approach. In scenarios where model predictions are uncertain or additional validation is necessary, these alerts could guide targeted laboratory measurements, optimizing resource use and improving public safety.

The performance of the current methodology may be even better than reported, as inaccuracies in the precise location of samples can result in misclassifying land instead of water areas. Improving the accuracy of sample geolocation is likely to enhance the performance of the trained model. Finally, while severity assessments at specific points within a water body are often desired, averaging predictions across multiple points could provide a more accurate representation of the overall severity for the entire or a specific part of the water body.

\section{Conclusions}
\label{sec:Conclusions}
The integration of artificial intelligence models with multiple free remote sensing data sources enables the rapid and accurate classification of algal bloom severity at any time and location, eliminating the need for in situ sampling. Platforms such as Google Earth Engine, Microsoft’s Planetary Computer, and NOAA provide quick access to essential data, significantly reducing the cost and time associated with downloading and processing raw data from other sources while still delivering reliable results. The color of the water body is crucial for assessing algal bloom severity at specific points, and using Sentinel-2 data as the sole optical satellite source proved adequate for this task. The combination of three distinct data sources—optical satellite imagery, climatological data, and elevation data—enhanced model performance, with temperature, rain, and wind emerging as the most significant climatological features. The NIR and two SWIR bands were the most important optical satellite bands while the altitude was the most important feature of the elevation data. While tree models alone were sufficient for the task, incorporating neural networks into the final solution improved both performance and reliability. The extensive geographic coverage of the dataset, encompassing the entire U.S., provided the necessary information for developing a high-performing modeling methodology, as data coming from a single city would likely lack the diversity required for effective model training. Trained models can be applied to estimate algal bloom severity across the U.S. quite accurately. Ultimately, a monitoring system centered around the presented methodology could serve as a cost-effective decision-support tool for identifying potential hazards, thereby enhancing public safety.

\section*{Declaration of Competing Interest}
The author declares that has no known competing financial interests or personal relationships that could have appeared to influence the work reported in this paper.

\section*{Data availability}
\label{sec:Data availability}
Data are available from sources as described in \autoref{sec:Data sources}. Metadata as well as train labels files are available for download at \href{https://www.drivendata.org/competitions/143/tick-tick-bloom/data/}{competition's website}.

\section*{Code availability}
\label{sec:Code availability} The complete running code as described in present research is available on \url{https://github.com/IoannisNasios/HarmfulAlgalBloomDetection}

\bibliographystyle{cas-model2-names}

\bibliography{refs}

\end{document}